\documentclass[acmtog]{acm}
\usepackage{color}
\usepackage{subfigure}
\usepackage{units}
\usepackage{expl3}
\usepackage{amsmath}
\usepackage{multirow}
\usepackage{colortbl}
\usepackage{algorithm}
\usepackage[noend]{algpseudocode}

\definecolor{Yellow}{rgb}{1,1, 0.6}
\definecolor{Red}{rgb}{1, 0.6, 0.6}
 
\usepackage{booktabs}
\setcitestyle{square}

\renewcommand\footnotetextcopyrightpermission[1]{}
\makeatletter
\renewcommand\@formatdoi[1]{\ignorespaces}
\makeatother

\begin{document}

\title{Vulcan Centaur: towards end-to-end real-time perception in lunar rovers}

\author{J. {de Curt\'o i D\'iAz} and R. A. Duvall.}
\affiliation{\institution{\\Iris Lunar Rover. Carnegie Mellon.}}

\renewcommand{\shortauthors}{De Curt\'o i D\'iAz and Duvall.}
\renewcommand{\shorttitle}{Vulcan Centaur: towards end-to-end real-time perception in lunar rovers}

\authorsaddresses{decurtoidiaz@ieee.org, rduvall@andrew.cmu.edu}

\begin{teaserfigure}
\centering
\includegraphics[width=\textwidth]{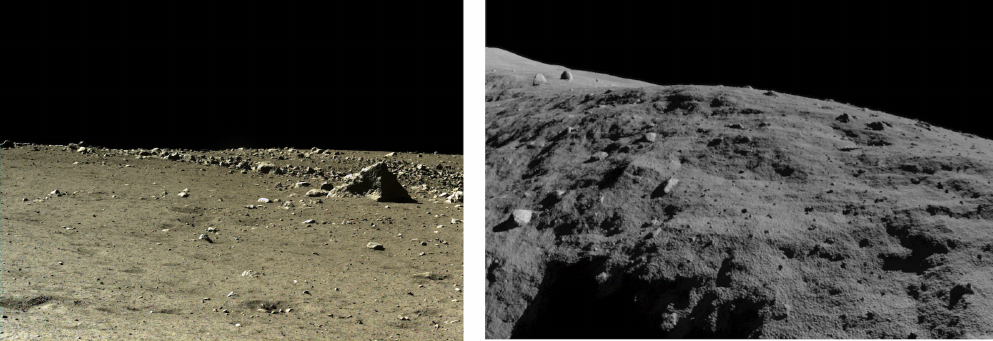}
   \caption{Left: Real image from the Moon. Right: Synthetic Moon.}
\label{fgr:c_3}
\end{teaserfigure}

\ExplSyntaxOn
\newcommand\latinabbrev[1]{
  \peek_meaning:NTF . {
    #1\@}%
  { \peek_catcode:NTF a {
      #1.\@ }%
    {#1.\@}}}
\ExplSyntaxOff

\def\eg{e.g. }
\def\etal{et al. }


\begin{abstract}
We introduce a new real-time pipeline for Simultaneous Localization and Mapping (SLAM) and Visual Inertial Odometry (VIO) in the context of planetary rovers. We leverage prior information of the location of the lander to propose an object-level SLAM approach that optimizes pose and shape of the lander together with camera trajectories of the rover. As a further refinement step, we propose to use techniques of interpolation between adjacent temporal samples; videlicet synthesizing non-existing images to improve the overall accuracy of the system. The experiments are conducted in the context of the Iris Lunar Rover, a nano-rover that will be deployed in lunar terrain in 2021 as the flagship of Carnegie Mellon, being the first unmanned rover of America to be on the Moon.
\end{abstract}

\maketitle
\fancyfoot{}
\thispagestyle{empty}


\section{Introduction}
\label{sn:introduction}

Our aim is to present a novel pipeline to deploy state-of-the-art DL techniques in planetary rovers. With the advent of a new wave of planetary exploration missions, the need to call on generalizable perception and control systems that can operate autonomously in other worlds will become ubiquitous in the coming years.


\section{Overall System}
\label{sn:os}

Following the design principles and the perception pipeline proposed in \cite{Allan19} in the context of the NASA Mission Resource Prospector, we put forward an improved technique for Visual Odometry (VIO) that could be exploited in a rover of the same characteristics. Although at the present time data from the Moon is scarce, there are already some open datasets available in analogue environments such as the POLAR Stereo Dataset \cite{Wong17} that includes stereo pairs and LiDAR information or \cite{Vayugundla18}, that contains IMU, stereo pairs and odometry plus some additional localization data, all obtained on Mount Etna. Specifically for the task of semantic segmentation, Kaggle provides images from a rendered environment of the Moon and masks. More recently, as a benchmark for tasks of Computer Vision in the context of space exploration, a dataset containing PNG images and positioning information from the mission Chang'E-4 to the Moon has been released \cite{DeCurtoandDuvall20}, the data from CE4 consists on post-processed original files from the mission Chang'E\footnote{\href{http://moon.bao.ac.cn/}{moon.bao.ac.cn}.}. Our specific sensor suite, that will be on-board the Iris Lunar Rover \cite{DeCurtoandDuvall20}, a project led by Carnegie Mellon that will deploy a four pound rover into the surface of the Moon by 2021 and that will be the first unmanned rover of America to explore the surface of the Moon, consists on IMU, two high-fidelity cameras and odometry sensors. Furthermore, it also has a UWB module \cite{Ledergerber15,Mueller15,Alarifi16,Xu20} on-board to localize the rover with respect to the lander.


\section{SLAM/VIO}
\label{sn:slam}

Simultaneous Localization and Mapping (SLAM) and Visual Inertial Odometry (VIO) are defined as a function that transform raw data from the sensors into a distribution over the states of the robot. SLAM and VIO \cite{Schneider18,Usenko19} have been for decades unparalleled problems in robot perception and state estimation. Although typical dense SLAM systems are not differentiable, new approaches to solve this problem propose gradient-based learning over computational graphs to go all the way from 3D maps to 2D pixels \cite{Murthy20}.
\\

The first task to tackle in geometric computer vision, being SLAM \cite{Newcombe11,Engel14,Engel17}, Structure-from-Motion (SfM) \cite{Snavely06,Agarwal09,Tateno17,Bloesch18,Teed20,Graham20}, camera calibration or image matching, is to extract interest points \cite{Ono18,DeTone18} from still images. We can define interest points as 2D specific locations in a given sample which can be considered stable and repeatable along different ambient conditions and viewpoints. The techniques used to traditionally attack this problem pertain to Multiple View Geometry \cite{Hartley03}, a subfield of mathematics that sets forth theorems and algorithms built on the assumption that those interest points can indeed be reliably extracted and matched across overlapping frames. Natheless, real-world computer vision operates on raw images that are far from the idealized conditions assumed in the proposed theory. Blending traditional modules with learning representations have lately been proven to be incredibly effective \cite{DeTone18,Tang19,Yang20} as a way to bridge the gap between the conditions that we face in the real world and the assumptions made to design the algorithms. Plentiful of approaches also explore unsupervised learning of depth and ego-motion \cite{Godard17,Zhou17,Yin18}.
\\

State-of-the-art approaches also deal with related problems such as object-level SLAM, that is, a system capable of optimizing object poses and shapes together with camera trajectory \cite{Suenderhauf17,McCormac18,Sucar20}. Although a SLAM system capable of incrementally mapping multi-object scenes seems not related to our task, its importance is revealed when we understand the fact that in many occasions the rover will localize itself with respect to the lander, which location is known; therefore a SLAM solution capable of optimizing the pose and shape of the lander along camera trajectory of the rover, would be distinctly adequate. With respect to this, we have to bear in mind that the principal technique that the rover will be using on-board to localize itself will be the UWB module \cite{Xu20}; that will indeed use the lander as a way-station for data communication. The reason for this is that critical weight and power can be hugely saved using RF for communication and state estimation. Thus, SLAM and VIO computation will be done on-ground. Using the same philosophy, it seems natural also to rely on a technique that will jointly optimize pose and shape of the lander together with camera trajectories.


\section{Shape and Pose of the Lander}
\label{sn:lander}

\begin{figure*}
\centering
   \includegraphics[scale=0.14]{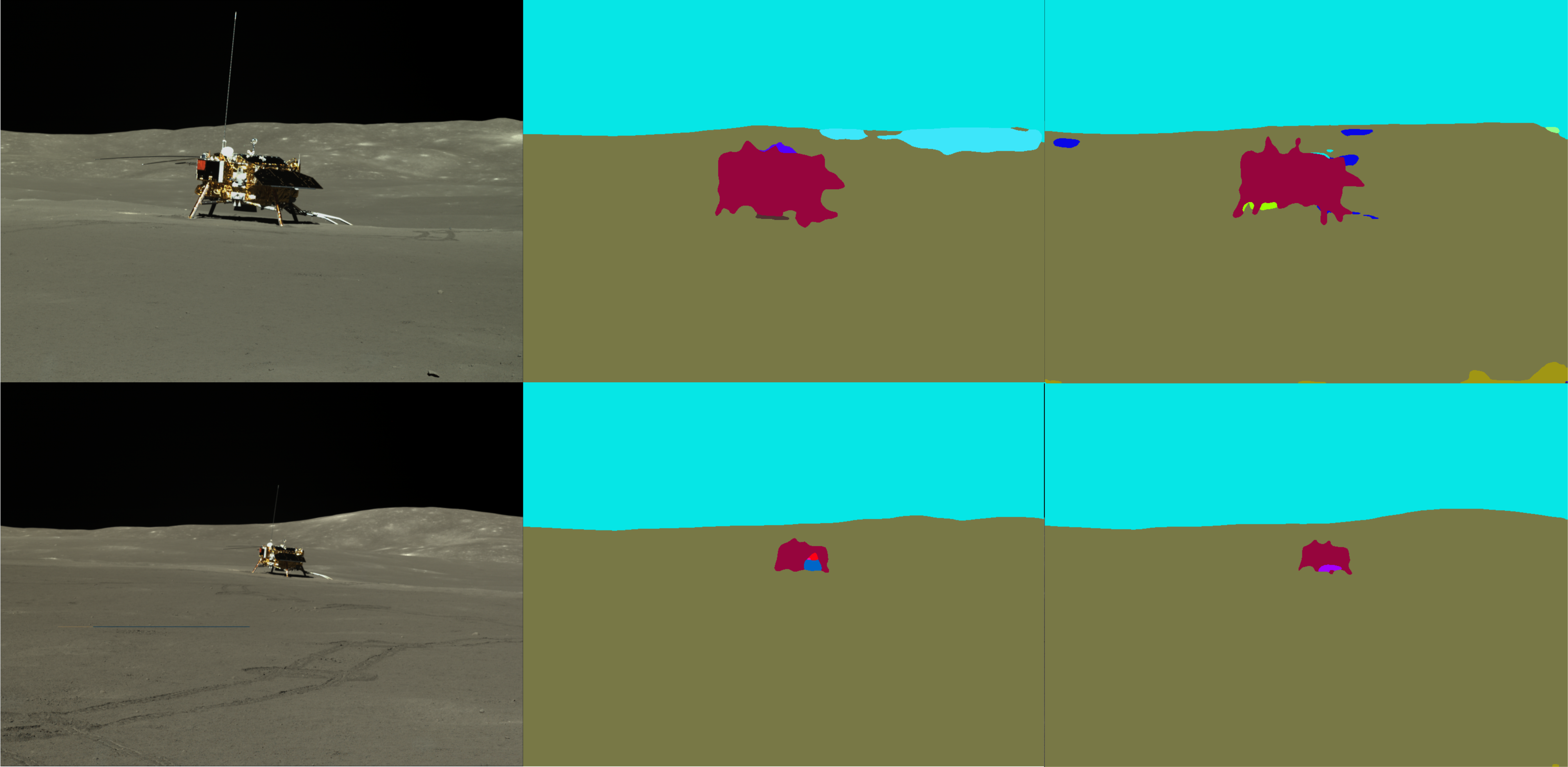}
   \caption{\textbf{Segmentation of the Lander}. Left: Image from CE4 \cite{DeCurtoandDuvall20}. In particular we are using color images from the panoramic camera of the rover of the mission to the Moon Chang'E-4. Middle: Generated mask given by model DilatedResnet-101 \cite{Zhao17,Zhou18}. Right: Generated mask given by model UperNet-101 \cite{Lin17,Xiao18}.}
\label{fgr:segmentation_ce4}
\end{figure*}

We assume here that we have a segmentation mask of the lander that in our specific case is obtained by the use of semantic segmentation \cite{Arbelaez11,Arbelaez14,Hariharan15,Mostajabi15,Yu16, Chen17,Chen17_2, Chen18, Chen18_3,He17,Zhou17}. On some of these approaches, the segmentation process is guided by the use of a prior object detector \cite{Girshick15,Ren15,Redmon16,Liu16,Redmon17,Huang17}. Specifically, we finetune our model building on DilatedResnet-101 \cite{Zhao17,Zhou18} and UperNet-101 \cite{Lin17,Xiao18} trained on ADE20K \cite{Zhou17}. Some examples of the mask given by our segmenter can be observed in Figure \ref{fgr:segmentation_ce4}. To infer the shape and pose we will leverage existing techniques \cite{Sucar20} that given a depth image, full shape and pose is determined. These techniques normally address multi-object categories; where a previous classification step and object observation is necessary, however our approach is somewhat simpler in the sense that the only object under consideration will be the lander per se.

\section{Temporal Interpolation between Subsequent Samples}
\label{sn:interpolation}

In the absence of continuous data between adjacent temporal samples given by the camera and to mitigate the effects that this will incur in the algorithms used to localize the rover, we propose to adopt techniques from video frame interpolation. Although signaling breakthroughs have been achieved by the use of recent deep convolutional neural networks, the quality of the resulting samples is often dubious due to object motion or occlusions. The main aim here is to synthesize non-existent frames in-between original samples to improve accuracy in the proposed VIO/SLAM approaches. Specifically for this purpose, we build on a recent depth-aware flow projection layer that achieves compelling upshots to synthesize intermediate sequences \cite{Bao19}.


\bibliographystyle{ACM-Reference-Format}

\begin{thebibliography}{52}


\ifx \showCODEN    \undefined \def \showCODEN     #1{\unskip}     \fi
\ifx \showDOI      \undefined \def \showDOI       #1{#1}\fi
\ifx \showISBNx    \undefined \def \showISBNx     #1{\unskip}     \fi
\ifx \showISBNxiii \undefined \def \showISBNxiii  #1{\unskip}     \fi
\ifx \showISSN     \undefined \def \showISSN      #1{\unskip}     \fi
\ifx \showLCCN     \undefined \def \showLCCN      #1{\unskip}     \fi
\ifx \shownote     \undefined \def \shownote      #1{#1}          \fi
\ifx \showarticletitle \undefined \def \showarticletitle #1{#1}   \fi
\ifx \showURL      \undefined \def \showURL       {\relax}        \fi
\providecommand\bibfield[2]{#2}
\providecommand\bibinfo[2]{#2}
\providecommand\natexlab[1]{#1}
\providecommand\showeprint[2][]{arXiv:#2}

\bibitem[\protect\citeauthoryear{Agarwal, Snavely, Simon, Seitz, and
  Szeliski}{Agarwal et~al\mbox{.}}{2009}]%
        {Agarwal09}
\bibfield{author}{\bibinfo{person}{S. Agarwal}, \bibinfo{person}{N. Snavely},
  \bibinfo{person}{I. Simon}, \bibinfo{person}{S.~M. Seitz}, {and}
  \bibinfo{person}{R. Szeliski}.} \bibinfo{year}{2009}\natexlab{}.
\newblock \showarticletitle{Building Rome on a Day}.
\newblock \bibinfo{journal}{\emph{ICCV}} (\bibinfo{year}{2009}).
\newblock


\bibitem[\protect\citeauthoryear{Alarifi, Al-Salman, Alsaleh, Alnafessah,
  Al-Hadhrami, Al-Ammar, and Al-Khalifa}{Alarifi et~al\mbox{.}}{2016}]%
        {Alarifi16}
\bibfield{author}{\bibinfo{person}{A. Alarifi}, \bibinfo{person}{A. Al-Salman},
  \bibinfo{person}{M. Alsaleh}, \bibinfo{person}{A. Alnafessah},
  \bibinfo{person}{S. Al-Hadhrami}, \bibinfo{person}{M.~A. Al-Ammar}, {and}
  \bibinfo{person}{H.~S. Al-Khalifa}.} \bibinfo{year}{2016}\natexlab{}.
\newblock \showarticletitle{Ultra Wideband Indoor Positioning Technologies:
  Analysis and Recent Advances}.
\newblock \bibinfo{journal}{\emph{Sensors}} (\bibinfo{year}{2016}).
\newblock


\bibitem[\protect\citeauthoryear{Allan, Wong, Furlong, Rogg, McMichael, Welsh,
  Chen, Peters, Gerkey, Quigley, Shirley, Deans, Cannon, and Fong}{Allan
  et~al\mbox{.}}{2019}]%
        {Allan19}
\bibfield{author}{\bibinfo{person}{M. Allan}, \bibinfo{person}{U. Wong},
  \bibinfo{person}{P.~M. Furlong}, \bibinfo{person}{A. Rogg},
  \bibinfo{person}{S. McMichael}, \bibinfo{person}{T. Welsh},
  \bibinfo{person}{I. Chen}, \bibinfo{person}{S. Peters}, \bibinfo{person}{B.
  Gerkey}, \bibinfo{person}{M. Quigley}, \bibinfo{person}{M. Shirley},
  \bibinfo{person}{M. Deans}, \bibinfo{person}{H. Cannon}, {and}
  \bibinfo{person}{T. Fong}.} \bibinfo{year}{2019}\natexlab{}.
\newblock \showarticletitle{Planetary Rover Simulation for Lunar Exploration
  Missions}.
\newblock \bibinfo{journal}{\emph{IEEE Aerospace Conference}}
  (\bibinfo{year}{2019}).
\newblock


\bibitem[\protect\citeauthoryear{Arbel\'aez, Maire, Fowlkes, and
  Malik}{Arbel\'aez et~al\mbox{.}}{2011}]%
        {Arbelaez11}
\bibfield{author}{\bibinfo{person}{P. Arbel\'aez}, \bibinfo{person}{M. Maire},
  \bibinfo{person}{C. Fowlkes}, {and} \bibinfo{person}{J. Malik}.}
  \bibinfo{year}{2011}\natexlab{}.
\newblock \showarticletitle{Contour Detection and Hierarchical Image
  Segmentation}.
\newblock \bibinfo{journal}{\emph{TPAMI}} (\bibinfo{year}{2011}).
\newblock


\bibitem[\protect\citeauthoryear{Arbel\'aez, Pont-Tuset, Barron, Marques, and
  Malik}{Arbel\'aez et~al\mbox{.}}{2014}]%
        {Arbelaez14}
\bibfield{author}{\bibinfo{person}{P. Arbel\'aez}, \bibinfo{person}{J.
  Pont-Tuset}, \bibinfo{person}{J.~T. Barron}, \bibinfo{person}{F. Marques},
  {and} \bibinfo{person}{J. Malik}.} \bibinfo{year}{2014}\natexlab{}.
\newblock \showarticletitle{Multiscale Combinatorial Grouping}.
\newblock \bibinfo{journal}{\emph{CVPR}} (\bibinfo{year}{2014}).
\newblock


\bibitem[\protect\citeauthoryear{Bao, Lai, Ma, Zhang, Gao, and Yang}{Bao
  et~al\mbox{.}}{2019}]%
        {Bao19}
\bibfield{author}{\bibinfo{person}{W. Bao}, \bibinfo{person}{W.-S. Lai},
  \bibinfo{person}{C. Ma}, \bibinfo{person}{X. Zhang}, \bibinfo{person}{Z.
  Gao}, {and} \bibinfo{person}{M.-H. Yang}.} \bibinfo{year}{2019}\natexlab{}.
\newblock \showarticletitle{Depth-Aware Video Frame Interpolation}.
\newblock \bibinfo{journal}{\emph{CVPR}} (\bibinfo{year}{2019}).
\newblock


\bibitem[\protect\citeauthoryear{Bloesch, Czarnowski, Clark, Leutenegger, and
  Davison}{Bloesch et~al\mbox{.}}{2018}]%
        {Bloesch18}
\bibfield{author}{\bibinfo{person}{M. Bloesch}, \bibinfo{person}{J.
  Czarnowski}, \bibinfo{person}{R. Clark}, \bibinfo{person}{S. Leutenegger},
  {and} \bibinfo{person}{A. Davison}.} \bibinfo{year}{2018}\natexlab{}.
\newblock \showarticletitle{CodeSLAM - Learning a Compact, Optimisable
  Representation for Dense Visual SLAM}.
\newblock \bibinfo{journal}{\emph{CVPR}} (\bibinfo{year}{2018}).
\newblock


\bibitem[\protect\citeauthoryear{Chen, Hermans, Papandreou, Schroff, Wang, and
  Adam}{Chen et~al\mbox{.}}{2018a}]%
        {Chen18_3}
\bibfield{author}{\bibinfo{person}{L. Chen}, \bibinfo{person}{A. Hermans},
  \bibinfo{person}{G. Papandreou}, \bibinfo{person}{F. Schroff},
  \bibinfo{person}{P. Wang}, {and} \bibinfo{person}{H. Adam}.}
  \bibinfo{year}{2018}\natexlab{a}.
\newblock \showarticletitle{{MaskLab}: Instance segmentation by refining object
  detection with semantic and direction features}.
\newblock \bibinfo{journal}{\emph{CVPR}} (\bibinfo{year}{2018}).
\newblock


\bibitem[\protect\citeauthoryear{Chen, Papandreou, Kokkinos, Murphy, and
  Yuille}{Chen et~al\mbox{.}}{2017a}]%
        {Chen17}
\bibfield{author}{\bibinfo{person}{L. Chen}, \bibinfo{person}{G. Papandreou},
  \bibinfo{person}{I. Kokkinos}, \bibinfo{person}{K. Murphy}, {and}
  \bibinfo{person}{A.~L. Yuille}.} \bibinfo{year}{2017}\natexlab{a}.
\newblock \showarticletitle{{DeepLab}: Semantic Image Segmentation with Deep
  Convolutional Nets, Atrous Convolution, and Fully Connected CRFs}.
\newblock \bibinfo{journal}{\emph{TPAMI}} (\bibinfo{year}{2017}).
\newblock


\bibitem[\protect\citeauthoryear{Chen, Papandreou, Schroff, and Adam}{Chen
  et~al\mbox{.}}{2017b}]%
        {Chen17_2}
\bibfield{author}{\bibinfo{person}{L. Chen}, \bibinfo{person}{G. Papandreou},
  \bibinfo{person}{F. Schroff}, {and} \bibinfo{person}{H. Adam}.}
  \bibinfo{year}{2017}\natexlab{b}.
\newblock \showarticletitle{Rethinking Atrous Convolution for Semantic Image
  Segmentation}.
\newblock \bibinfo{journal}{\emph{arXiv:1706.05587}} (\bibinfo{year}{2017}).
\newblock


\bibitem[\protect\citeauthoryear{Chen, Zhu, Papandreou, Schroff, and Adam}{Chen
  et~al\mbox{.}}{2018b}]%
        {Chen18}
\bibfield{author}{\bibinfo{person}{L. Chen}, \bibinfo{person}{Y. Zhu},
  \bibinfo{person}{G. Papandreou}, \bibinfo{person}{F. Schroff}, {and}
  \bibinfo{person}{H. Adam}.} \bibinfo{year}{2018}\natexlab{b}.
\newblock \showarticletitle{Encoder-Decoder with Atrous Separable Convolution
  for Semantic Image Segmentation}.
\newblock \bibinfo{journal}{\emph{ECCV}} (\bibinfo{year}{2018}).
\newblock


\bibitem[\protect\citeauthoryear{{de Curt\'o} and Duvall}{{de Curt\'o} and
  Duvall}{2020}]%
        {DeCurtoandDuvall20}
\bibfield{author}{\bibinfo{person}{J. {de Curt\'o}} {and} \bibinfo{person}{R.
  Duvall}.} \bibinfo{year}{2020}\natexlab{}.
\newblock \showarticletitle{Cycle-consistent Generative Adversarial Networks
  for Neural Style Transfer using data from Chang'E-4}.
\newblock \bibinfo{journal}{\emph{arXiv:2011.11627}} (\bibinfo{year}{2020}).
\newblock


\bibitem[\protect\citeauthoryear{DeTone, Malisiewicz, and Rabinovich}{DeTone
  et~al\mbox{.}}{2018}]%
        {DeTone18}
\bibfield{author}{\bibinfo{person}{D. DeTone}, \bibinfo{person}{T.
  Malisiewicz}, {and} \bibinfo{person}{A. Rabinovich}.}
  \bibinfo{year}{2018}\natexlab{}.
\newblock \showarticletitle{{SuperPoint}: Self-Supervised Interest Point
  Detection and Description}.
\newblock \bibinfo{journal}{\emph{CVPR}} (\bibinfo{year}{2018}).
\newblock


\bibitem[\protect\citeauthoryear{Engel, Koltun, and Cremers}{Engel
  et~al\mbox{.}}{2017}]%
        {Engel17}
\bibfield{author}{\bibinfo{person}{J. Engel}, \bibinfo{person}{V. Koltun},
  {and} \bibinfo{person}{D. Cremers}.} \bibinfo{year}{2017}\natexlab{}.
\newblock \showarticletitle{Direct Sparse Odometry}.
\newblock \bibinfo{journal}{\emph{T-PAMI}} (\bibinfo{year}{2017}).
\newblock


\bibitem[\protect\citeauthoryear{Engel, Sch{\"o}ps, and Cremers}{Engel
  et~al\mbox{.}}{2014}]%
        {Engel14}
\bibfield{author}{\bibinfo{person}{J. Engel}, \bibinfo{person}{T. Sch{\"o}ps},
  {and} \bibinfo{person}{D. Cremers}.} \bibinfo{year}{2014}\natexlab{}.
\newblock \showarticletitle{{LSD-SLAM}: Large-Scale Direct Monocular SLAM}.
\newblock \bibinfo{journal}{\emph{ECCV}} (\bibinfo{year}{2014}).
\newblock


\bibitem[\protect\citeauthoryear{Girshick}{Girshick}{2015}]%
        {Girshick15}
\bibfield{author}{\bibinfo{person}{R. Girshick}.}
  \bibinfo{year}{2015}\natexlab{}.
\newblock \showarticletitle{FAST R-CNN}.
\newblock \bibinfo{journal}{\emph{ICCV}} (\bibinfo{year}{2015}).
\newblock


\bibitem[\protect\citeauthoryear{Godard, {Mac Aodha}, and Brostow}{Godard
  et~al\mbox{.}}{2017}]%
        {Godard17}
\bibfield{author}{\bibinfo{person}{C. Godard}, \bibinfo{person}{O. {Mac
  Aodha}}, {and} \bibinfo{person}{G. Brostow}.}
  \bibinfo{year}{2017}\natexlab{}.
\newblock \showarticletitle{Unsupervised Monocular Depth Estimation with
  Left-Right Consistency}.
\newblock \bibinfo{journal}{\emph{CVPR}} (\bibinfo{year}{2017}).
\newblock


\bibitem[\protect\citeauthoryear{Graham and Novotny}{Graham and
  Novotny}{2020}]%
        {Graham20}
\bibfield{author}{\bibinfo{person}{B. Graham} {and} \bibinfo{person}{D.
  Novotny}.} \bibinfo{year}{2020}\natexlab{}.
\newblock \showarticletitle{RidgeSfM: Structure from Motion via Robust Pairwise
  Matching Under Depth Uncertainty}.
\newblock \bibinfo{journal}{\emph{3DV}} (\bibinfo{year}{2020}).
\newblock


\bibitem[\protect\citeauthoryear{Hariharan, Arbel\'aez, Girshick, and
  Malik}{Hariharan et~al\mbox{.}}{2015}]%
        {Hariharan15}
\bibfield{author}{\bibinfo{person}{B. Hariharan}, \bibinfo{person}{P.
  Arbel\'aez}, \bibinfo{person}{R. Girshick}, {and} \bibinfo{person}{J.
  Malik}.} \bibinfo{year}{2015}\natexlab{}.
\newblock \showarticletitle{Hypercolumns for Object Segmentation and
  Fine-grained Localization}.
\newblock \bibinfo{journal}{\emph{CVPR}} (\bibinfo{year}{2015}).
\newblock


\bibitem[\protect\citeauthoryear{Hartley and Zisserman}{Hartley and
  Zisserman}{2003}]%
        {Hartley03}
\bibfield{author}{\bibinfo{person}{R. Hartley} {and} \bibinfo{person}{A.
  Zisserman}.} \bibinfo{year}{2003}\natexlab{}.
\newblock \bibinfo{booktitle}{\emph{Multiple View Geometry in Computer
  Vision}}.
\newblock


\bibitem[\protect\citeauthoryear{He, Gkioxari, Doll\'ar, and Girshick}{He
  et~al\mbox{.}}{2017}]%
        {He17}
\bibfield{author}{\bibinfo{person}{K. He}, \bibinfo{person}{G. Gkioxari},
  \bibinfo{person}{P. Doll\'ar}, {and} \bibinfo{person}{R. Girshick}.}
  \bibinfo{year}{2017}\natexlab{}.
\newblock \showarticletitle{MASK R-CNN}.
\newblock \bibinfo{journal}{\emph{ICCV}} (\bibinfo{year}{2017}).
\newblock


\bibitem[\protect\citeauthoryear{Huang, Rathod, Sun, Zhu, Korattikara, Fathi,
  Fischer, Wojna, Song, Guadarrama, and Murphy}{Huang et~al\mbox{.}}{2017}]%
        {Huang17}
\bibfield{author}{\bibinfo{person}{J. Huang}, \bibinfo{person}{V. Rathod},
  \bibinfo{person}{C. Sun}, \bibinfo{person}{M. Zhu}, \bibinfo{person}{A.
  Korattikara}, \bibinfo{person}{A. Fathi}, \bibinfo{person}{I. Fischer},
  \bibinfo{person}{Z. Wojna}, \bibinfo{person}{Y. Song}, \bibinfo{person}{S.
  Guadarrama}, {and} \bibinfo{person}{K. Murphy}.}
  \bibinfo{year}{2017}\natexlab{}.
\newblock \showarticletitle{Speed/accuracy trade-offs for modern convolutional
  object detectors}.
\newblock \bibinfo{journal}{\emph{CVPR}} (\bibinfo{year}{2017}).
\newblock


\bibitem[\protect\citeauthoryear{Ledergerber, Hamer, and D'Andrea}{Ledergerber
  et~al\mbox{.}}{2015}]%
        {Ledergerber15}
\bibfield{author}{\bibinfo{person}{A. Ledergerber}, \bibinfo{person}{M. Hamer},
  {and} \bibinfo{person}{R. D'Andrea}.} \bibinfo{year}{2015}\natexlab{}.
\newblock \showarticletitle{A Robot Self-Localization System using One-Way
  Ultra-Wideband Communication}.
\newblock \bibinfo{journal}{\emph{IROS}} (\bibinfo{year}{2015}).
\newblock


\bibitem[\protect\citeauthoryear{Lin, Doll\'ar, Girshick, He, Hariharan, and
  Belongie}{Lin et~al\mbox{.}}{2017}]%
        {Lin17}
\bibfield{author}{\bibinfo{person}{T.-Y. Lin}, \bibinfo{person}{P. Doll\'ar},
  \bibinfo{person}{R. Girshick}, \bibinfo{person}{K. He}, \bibinfo{person}{B.
  Hariharan}, {and} \bibinfo{person}{S. Belongie}.}
  \bibinfo{year}{2017}\natexlab{}.
\newblock \showarticletitle{Feature Pyramid Networks for Object Detection}.
\newblock \bibinfo{journal}{\emph{CVPR}} (\bibinfo{year}{2017}).
\newblock


\bibitem[\protect\citeauthoryear{Liu, Anguelov, Erhan, Szegedy, Reed, Fu, and
  Berg}{Liu et~al\mbox{.}}{2016}]%
        {Liu16}
\bibfield{author}{\bibinfo{person}{W. Liu}, \bibinfo{person}{D. Anguelov},
  \bibinfo{person}{D. Erhan}, \bibinfo{person}{C. Szegedy}, \bibinfo{person}{S.
  Reed}, \bibinfo{person}{C. Fu}, {and} \bibinfo{person}{A.~C. Berg}.}
  \bibinfo{year}{2016}\natexlab{}.
\newblock \showarticletitle{{SSD}: Single Shot MultiBox Detector}.
\newblock \bibinfo{journal}{\emph{ECCV}} (\bibinfo{year}{2016}).
\newblock


\bibitem[\protect\citeauthoryear{McCormac, Clark, Bloesch, Davison, and
  Leutenegger}{McCormac et~al\mbox{.}}{2018}]%
        {McCormac18}
\bibfield{author}{\bibinfo{person}{J. McCormac}, \bibinfo{person}{R. Clark},
  \bibinfo{person}{M. Bloesch}, \bibinfo{person}{A. Davison}, {and}
  \bibinfo{person}{S. Leutenegger}.} \bibinfo{year}{2018}\natexlab{}.
\newblock \showarticletitle{Fusion++: Volumetric Object-Level SLAM}.
\newblock \bibinfo{journal}{\emph{3DV}} (\bibinfo{year}{2018}).
\newblock


\bibitem[\protect\citeauthoryear{Mostajabi, Yadollahpour, and
  Shakhnarovich}{Mostajabi et~al\mbox{.}}{2015}]%
        {Mostajabi15}
\bibfield{author}{\bibinfo{person}{M. Mostajabi}, \bibinfo{person}{P.
  Yadollahpour}, {and} \bibinfo{person}{G. Shakhnarovich}.}
  \bibinfo{year}{2015}\natexlab{}.
\newblock \showarticletitle{Feedforward semantic segmentation with zoom-out
  features}.
\newblock \bibinfo{journal}{\emph{CVPR}} (\bibinfo{year}{2015}).
\newblock


\bibitem[\protect\citeauthoryear{Mueller, Hamer, and D'Andrea}{Mueller
  et~al\mbox{.}}{2015}]%
        {Mueller15}
\bibfield{author}{\bibinfo{person}{M.~W. Mueller}, \bibinfo{person}{M. Hamer},
  {and} \bibinfo{person}{R. D'Andrea}.} \bibinfo{year}{2015}\natexlab{}.
\newblock \showarticletitle{Fusing Ultra-Wideband Range Measurements with
  Accelerometers and Rate Gyroscopes for Quadrocopter State Estimation}.
\newblock \bibinfo{journal}{\emph{ICRA}} (\bibinfo{year}{2015}).
\newblock


\bibitem[\protect\citeauthoryear{Murthy, Saryazdi, Iyer, and Paull}{Murthy
  et~al\mbox{.}}{2020}]%
        {Murthy20}
\bibfield{author}{\bibinfo{person}{K. Murthy}, \bibinfo{person}{S. Saryazdi},
  \bibinfo{person}{G. Iyer}, {and} \bibinfo{person}{L. Paull}.}
  \bibinfo{year}{2020}\natexlab{}.
\newblock \showarticletitle{gradSLAM: Dense SLAM meets automatic
  differentiation}.
\newblock \bibinfo{journal}{\emph{ICRA}} (\bibinfo{year}{2020}).
\newblock


\bibitem[\protect\citeauthoryear{Newcombe, Lovegrove, and Davison}{Newcombe
  et~al\mbox{.}}{2011}]%
        {Newcombe11}
\bibfield{author}{\bibinfo{person}{R. Newcombe}, \bibinfo{person}{S.
  Lovegrove}, {and} \bibinfo{person}{A. Davison}.}
  \bibinfo{year}{2011}\natexlab{}.
\newblock \showarticletitle{DTAM: Dense tracking and mapping in real-time}.
\newblock \bibinfo{journal}{\emph{ICCV}} (\bibinfo{year}{2011}).
\newblock


\bibitem[\protect\citeauthoryear{Ono, Trulls, Fua, and Yi}{Ono
  et~al\mbox{.}}{2018}]%
        {Ono18}
\bibfield{author}{\bibinfo{person}{Y. Ono}, \bibinfo{person}{E. Trulls},
  \bibinfo{person}{P. Fua}, {and} \bibinfo{person}{K.~M. Yi}.}
  \bibinfo{year}{2018}\natexlab{}.
\newblock \showarticletitle{{LF-Net}: Learning Local Features from Images}.
\newblock \bibinfo{journal}{\emph{NIPS}} (\bibinfo{year}{2018}).
\newblock


\bibitem[\protect\citeauthoryear{Redmon, Divvala, Girshick, and Farhadi}{Redmon
  et~al\mbox{.}}{2016}]%
        {Redmon16}
\bibfield{author}{\bibinfo{person}{J. Redmon}, \bibinfo{person}{S. Divvala},
  \bibinfo{person}{R. Girshick}, {and} \bibinfo{person}{A. Farhadi}.}
  \bibinfo{year}{2016}\natexlab{}.
\newblock \showarticletitle{You Only Look Once: Unified, Real-Time Object
  Detection}.
\newblock \bibinfo{journal}{\emph{CVPR}} (\bibinfo{year}{2016}).
\newblock


\bibitem[\protect\citeauthoryear{Redmon and Farhadi}{Redmon and
  Farhadi}{2017}]%
        {Redmon17}
\bibfield{author}{\bibinfo{person}{J. Redmon} {and} \bibinfo{person}{A.
  Farhadi}.} \bibinfo{year}{2017}\natexlab{}.
\newblock \showarticletitle{YOLO9000: Better, Faster, Stronger}.
\newblock \bibinfo{journal}{\emph{CVPR}} (\bibinfo{year}{2017}).
\newblock


\bibitem[\protect\citeauthoryear{Ren, He, Girshick, and Sun}{Ren
  et~al\mbox{.}}{2015}]%
        {Ren15}
\bibfield{author}{\bibinfo{person}{S. Ren}, \bibinfo{person}{K. He},
  \bibinfo{person}{R. Girshick}, {and} \bibinfo{person}{J. Sun}.}
  \bibinfo{year}{2015}\natexlab{}.
\newblock \showarticletitle{FASTER R-CNN: Towards Real-Time Object Detection
  with Region Proposal Networks}.
\newblock \bibinfo{journal}{\emph{NIPS}} (\bibinfo{year}{2015}).
\newblock


\bibitem[\protect\citeauthoryear{Schneider, Dymczyk, Fehr, Egger, Lynen,
  Gilitschenski, and Siegwart}{Schneider et~al\mbox{.}}{2018}]%
        {Schneider18}
\bibfield{author}{\bibinfo{person}{T. Schneider}, \bibinfo{person}{M. Dymczyk},
  \bibinfo{person}{M. Fehr}, \bibinfo{person}{K. Egger}, \bibinfo{person}{S.
  Lynen}, \bibinfo{person}{I. Gilitschenski}, {and} \bibinfo{person}{R.
  Siegwart}.} \bibinfo{year}{2018}\natexlab{}.
\newblock \showarticletitle{maplab: An Open Framework for Research in
  Visual-inertial Mapping and Localization}.
\newblock \bibinfo{journal}{\emph{ICRA}} (\bibinfo{year}{2018}).
\newblock


\bibitem[\protect\citeauthoryear{Snavely, Seitz, and Szeliski}{Snavely
  et~al\mbox{.}}{2006}]%
        {Snavely06}
\bibfield{author}{\bibinfo{person}{N. Snavely}, \bibinfo{person}{S.~M. Seitz},
  {and} \bibinfo{person}{R. Szeliski}.} \bibinfo{year}{2006}\natexlab{}.
\newblock \showarticletitle{Photo Tourism: Exploring Photo Collections in 3D}.
\newblock \bibinfo{journal}{\emph{SIGGRAPH}} (\bibinfo{year}{2006}).
\newblock


\bibitem[\protect\citeauthoryear{Sucar, Wada, and Davison}{Sucar
  et~al\mbox{.}}{2020}]%
        {Sucar20}
\bibfield{author}{\bibinfo{person}{E. Sucar}, \bibinfo{person}{K. Wada}, {and}
  \bibinfo{person}{A. Davison}.} \bibinfo{year}{2020}\natexlab{}.
\newblock \showarticletitle{NodeSLAM: Neural Object Descriptors for Multi-View
  Shape Reconstruction}.
\newblock \bibinfo{journal}{\emph{3DV}} (\bibinfo{year}{2020}).
\newblock


\bibitem[\protect\citeauthoryear{S{\"u}nderhauf, Pham, Latif, Milford, and
  Reid}{S{\"u}nderhauf et~al\mbox{.}}{2017}]%
        {Suenderhauf17}
\bibfield{author}{\bibinfo{person}{N. S{\"u}nderhauf}, \bibinfo{person}{T.~T.
  Pham}, \bibinfo{person}{Y. Latif}, \bibinfo{person}{M. Milford}, {and}
  \bibinfo{person}{I. Reid}.} \bibinfo{year}{2017}\natexlab{}.
\newblock \showarticletitle{Meaningful Maps With Object-Oriented Semantic
  Mapping}.
\newblock \bibinfo{journal}{\emph{IROS}} (\bibinfo{year}{2017}).
\newblock


\bibitem[\protect\citeauthoryear{Tang and Tan}{Tang and Tan}{2019}]%
        {Tang19}
\bibfield{author}{\bibinfo{person}{C. Tang} {and} \bibinfo{person}{P. Tan}.}
  \bibinfo{year}{2019}\natexlab{}.
\newblock \showarticletitle{{BA-Net}: Dense Bundle Adjustment Network}.
\newblock \bibinfo{journal}{\emph{ICLR}} (\bibinfo{year}{2019}).
\newblock


\bibitem[\protect\citeauthoryear{Tateno, Tombari, Laina, and Navab}{Tateno
  et~al\mbox{.}}{2017}]%
        {Tateno17}
\bibfield{author}{\bibinfo{person}{K. Tateno}, \bibinfo{person}{F. Tombari},
  \bibinfo{person}{I. Laina}, {and} \bibinfo{person}{N. Navab}.}
  \bibinfo{year}{2017}\natexlab{}.
\newblock \showarticletitle{{CNN-SLAM}: Real-time dense monocular SLAM with
  learned depth prediction}.
\newblock \bibinfo{journal}{\emph{CVPR}} (\bibinfo{year}{2017}).
\newblock


\bibitem[\protect\citeauthoryear{Usenko, Demmel, Schubert, St{\"u}ckler, and
  Cremers}{Usenko et~al\mbox{.}}{2019}]%
        {Usenko19}
\bibfield{author}{\bibinfo{person}{V. Usenko}, \bibinfo{person}{N. Demmel},
  \bibinfo{person}{D. Schubert}, \bibinfo{person}{J. St{\"u}ckler}, {and}
  \bibinfo{person}{D. Cremers}.} \bibinfo{year}{2019}\natexlab{}.
\newblock \showarticletitle{Visual-Inertial Mapping with Non-Linear Factor
  Recovery}.
\newblock \bibinfo{journal}{\emph{{IEEE} Robotics and Automation Letters}}
  (\bibinfo{year}{2019}).
\newblock


\bibitem[\protect\citeauthoryear{Vayugundla, Steidle, Smisek, Schuster,
  Bussmann, and Wedler}{Vayugundla et~al\mbox{.}}{2018}]%
        {Vayugundla18}
\bibfield{author}{\bibinfo{person}{M. Vayugundla}, \bibinfo{person}{F.
  Steidle}, \bibinfo{person}{M. Smisek}, \bibinfo{person}{M.~J. Schuster},
  \bibinfo{person}{K. Bussmann}, {and} \bibinfo{person}{A. Wedler}.}
  \bibinfo{year}{2018}\natexlab{}.
\newblock \showarticletitle{Datasets of Long Range Navigation Experiments in a
  Moon Analogue Environment on Mount Etna}.
\newblock \bibinfo{journal}{\emph{International Symposium on Robotics}}
  (\bibinfo{year}{2018}).
\newblock


\bibitem[\protect\citeauthoryear{Wong, Nefian, Edwards, Buoyssounouse, Furlong,
  Deans, and Fong}{Wong et~al\mbox{.}}{2017}]%
        {Wong17}
\bibfield{author}{\bibinfo{person}{U. Wong}, \bibinfo{person}{A. Nefian},
  \bibinfo{person}{L. Edwards}, \bibinfo{person}{X. Buoyssounouse},
  \bibinfo{person}{P.~M. Furlong}, \bibinfo{person}{M. Deans}, {and}
  \bibinfo{person}{T. Fong}.} \bibinfo{year}{2017}\natexlab{}.
\newblock \showarticletitle{Polar Optical Lunar Analog Reconstruction (POLAR)
  Stereo Dataset}.
\newblock \bibinfo{journal}{\emph{NASA Ames Research Center}}
  (\bibinfo{year}{2017}).
\newblock


\bibitem[\protect\citeauthoryear{Xiao, Liu, Zhou, Jiang, and Sun}{Xiao
  et~al\mbox{.}}{2018}]%
        {Xiao18}
\bibfield{author}{\bibinfo{person}{T. Xiao}, \bibinfo{person}{Y. Liu},
  \bibinfo{person}{B. Zhou}, \bibinfo{person}{Y. Jiang}, {and}
  \bibinfo{person}{J. Sun}.} \bibinfo{year}{2018}\natexlab{}.
\newblock \showarticletitle{Unified Perceptual Parsing for Scene
  Understanding}.
\newblock \bibinfo{journal}{\emph{ECCV}} (\bibinfo{year}{2018}).
\newblock


\bibitem[\protect\citeauthoryear{Xu, Wang, Zhang, Qiu, and Shen}{Xu
  et~al\mbox{.}}{2020}]%
        {Xu20}
\bibfield{author}{\bibinfo{person}{H. Xu}, \bibinfo{person}{L. Wang},
  \bibinfo{person}{Y. Zhang}, \bibinfo{person}{K. Qiu}, {and}
  \bibinfo{person}{S. Shen}.} \bibinfo{year}{2020}\natexlab{}.
\newblock \showarticletitle{Decentralized Visual-Inertial-UWB Fusion for
  Relative State Estimation of Aerial Swarm}.
\newblock \bibinfo{journal}{\emph{ICRA}} (\bibinfo{year}{2020}).
\newblock


\bibitem[\protect\citeauthoryear{Yang, {von Stumberg}, Wang, and Cremers}{Yang
  et~al\mbox{.}}{2020}]%
        {Yang20}
\bibfield{author}{\bibinfo{person}{N. Yang}, \bibinfo{person}{L. {von
  Stumberg}}, \bibinfo{person}{R. Wang}, {and} \bibinfo{person}{D. Cremers}.}
  \bibinfo{year}{2020}\natexlab{}.
\newblock \showarticletitle{{D3VO}: Deep Depth, Deep Pose and Deep Uncertainty
  for Monocular Visual Odometry}.
\newblock \bibinfo{journal}{\emph{CVPR}} (\bibinfo{year}{2020}).
\newblock


\bibitem[\protect\citeauthoryear{Yin and Shi}{Yin and Shi}{2018}]%
        {Yin18}
\bibfield{author}{\bibinfo{person}{Z. Yin} {and} \bibinfo{person}{J. Shi}.}
  \bibinfo{year}{2018}\natexlab{}.
\newblock \showarticletitle{{GeoNet}: Unsupervised Learning of Dense Depth,
  Optical Flow and Camera Pose}.
\newblock \bibinfo{journal}{\emph{CVPR}} (\bibinfo{year}{2018}).
\newblock


\bibitem[\protect\citeauthoryear{Yu and Koltun}{Yu and Koltun}{2016}]%
        {Yu16}
\bibfield{author}{\bibinfo{person}{F. Yu} {and} \bibinfo{person}{V. Koltun}.}
  \bibinfo{year}{2016}\natexlab{}.
\newblock \showarticletitle{Multi-Scale Context Aggregation by Dilated
  Convolutions}.
\newblock \bibinfo{journal}{\emph{ICLR}} (\bibinfo{year}{2016}).
\newblock


\bibitem[\protect\citeauthoryear{Z.~Teed}{Z.~Teed}{2020}]%
        {Teed20}
\bibfield{author}{\bibinfo{person}{J.~Deng Z.~Teed}.}
  \bibinfo{year}{2020}\natexlab{}.
\newblock \showarticletitle{DeepV2D: Video to Depth with Differentiable
  Structure from Motion}.
\newblock \bibinfo{journal}{\emph{ICLR}} (\bibinfo{year}{2020}).
\newblock


\bibitem[\protect\citeauthoryear{Zhao, Shi, Qi, Wang, and Jia}{Zhao
  et~al\mbox{.}}{2017}]%
        {Zhao17}
\bibfield{author}{\bibinfo{person}{H. Zhao}, \bibinfo{person}{J. Shi},
  \bibinfo{person}{X. Qi}, \bibinfo{person}{X. Wang}, {and} \bibinfo{person}{J.
  Jia}.} \bibinfo{year}{2017}\natexlab{}.
\newblock \showarticletitle{Pyramid Scene Parsing Network}.
\newblock \bibinfo{journal}{\emph{CVPR}} (\bibinfo{year}{2017}).
\newblock


\bibitem[\protect\citeauthoryear{Zhou, Zhao, Puig, Xiao, Fidler, Barriuso, and
  Torralba}{Zhou et~al\mbox{.}}{2018}]%
        {Zhou18}
\bibfield{author}{\bibinfo{person}{B. Zhou}, \bibinfo{person}{H. Zhao},
  \bibinfo{person}{X. Puig}, \bibinfo{person}{T. Xiao}, \bibinfo{person}{S.
  Fidler}, \bibinfo{person}{A. Barriuso}, {and} \bibinfo{person}{A. Torralba}.}
  \bibinfo{year}{2018}\natexlab{}.
\newblock \showarticletitle{Semantic Understanding of Scenes through the ADE20K
  Dataset}.
\newblock \bibinfo{journal}{\emph{IJCV}} (\bibinfo{year}{2018}).
\newblock


\bibitem[\protect\citeauthoryear{Zhou, Brown, Snavely, and Lowe}{Zhou
  et~al\mbox{.}}{2017}]%
        {Zhou17}
\bibfield{author}{\bibinfo{person}{T. Zhou}, \bibinfo{person}{M. Brown},
  \bibinfo{person}{N. Snavely}, {and} \bibinfo{person}{D. Lowe}.}
  \bibinfo{year}{2017}\natexlab{}.
\newblock \showarticletitle{Unsupervised Learning of Depth and Ego-Motion from
  Video}.
\newblock \bibinfo{journal}{\emph{CVPR}} (\bibinfo{year}{2017}).
\newblock


\end{thebibliography}

\end{document}